\documentclass[letterpaper,10pt, conference]{IEEEtran}

\IEEEoverridecommandlockouts
                                          
\usepackage{graphicx}
\usepackage{cite}
\usepackage{amssymb}
\usepackage{bm}
\usepackage{tcolorbox}
\usepackage{amsmath}
\usepackage{tabularray}
\usepackage[super]{nth}
\usepackage{color,soul}
\usepackage{algorithm}  
\usepackage{algorithmic}
\usepackage{subcaption}
\usepackage{float}
\usepackage{fancyhdr}
\fancypagestyle{firstpage}{%
  \fancyhf{}
  
  \fancyhead[C]{\footnotesize This work has been submitted to the IEEE for possible publication. Copyright may be transferred without notice, after which this version may no longer be accessible.}
}

\title{\LARGE \bf
EKF-Based Depth Camera and Deep Learning Fusion for UAV-Person Distance Estimation and Following in SAR Operations
}

\author{
\IEEEauthorblockN{
Luka Šiktar\IEEEauthorrefmark{1},
Branimir Ćaran\IEEEauthorrefmark{1},
Bojan Šekoranja\IEEEauthorrefmark{1},
Marko Švaco\IEEEauthorrefmark{1}
}
\IEEEauthorblockA{
\IEEEauthorrefmark{1}
Faculty of Mechanical Engineering and Naval Architecture\\
University of Zagreb\\
Zagreb, Croatia\\
email: luka.siktar@fsb.hr
}
}

\begin{document}

\maketitle
\pagestyle{empty}
\thispagestyle{firstpage}

%%%%%%%%%%%%%%%%%%%%%%%%%%%%%%%%%%%%%%%%%%%%%%%%%%%%%%%%%%%%%%%%%%%%%%%%%%%%%%%%
\begin{abstract}

Vision-based Unmanned Aerial Vehicles (UAVs) frameworks aid human search tasks by detecting and recognizing specific individuals, then tracking and following them while maintaining a safe distance. A key safety requirement for UAV following is the accurate estimation of the distance between camera and target object under real-world conditions, achieved by fusing multiple image modalities. As part of the system for automatic people detection and face recognition using deep learning, in this paper we present the fusion of depth camera measurements and monocular camera-to-body distance estimation for robust tracking and following. Deep learning based filtering of depth camera data and estimation of camera-to-body distance from a monocular camera are achieved with YOLO-pose, enabling real-time fusion of depth information using the Extended Kalman Filter (EKF) algorithm. The proposed subsystem, designed for use in drones, estimates and measures the distance between the depth camera and the human body keypoints, to maintain the safe distance between the drone and the human target. Our system provides an accurate estimated distance, which has been validated against motion capture ground truth data. The system has been tested in real time indoors, where it reduces the average errors, RMSE and standard deviations of distance estimation up to 15,3\% in three tested scenarios. Based on the test results, the EKF fusion-based approach increases the depth detection range by reducing the errors outside the optimal depth camera working range. It also shows improved robustness and precision in challenging conditions, such as reflections and poor visibility, making it suitable for SAR.
\end{abstract}

\renewcommand\IEEEkeywordsname{Keywords}
\begin{IEEEkeywords}
\textit{UAV, SAR, Depth Estimation, Computer vision, Sensor fusion, Human detection and tracking, Visual tracking.}
\end{IEEEkeywords}

\section{Introduction}
\begin{figure}[h]
    \includegraphics[width=0.48\textwidth]{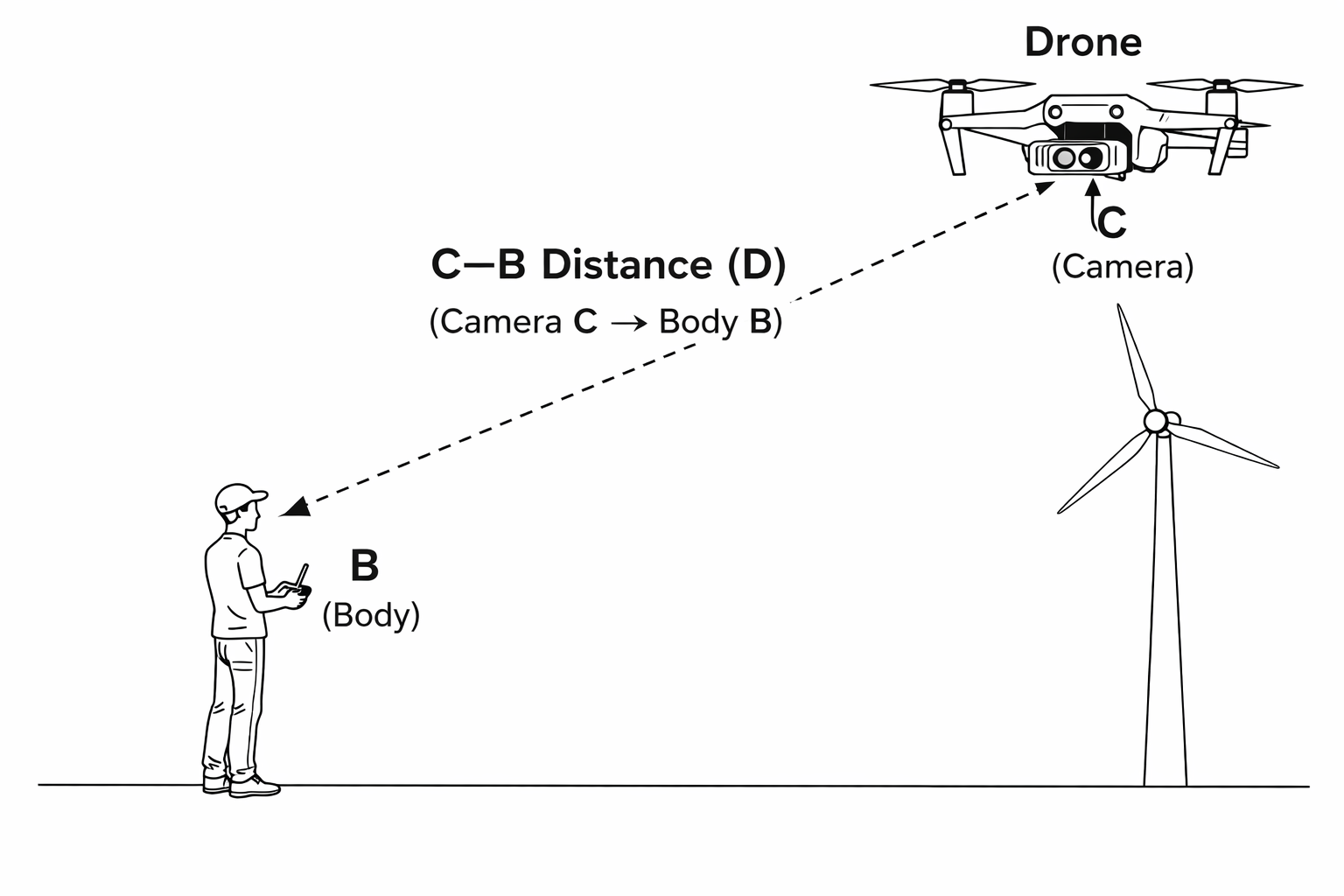}
    
    \caption{Camera-to-body (C-B) distance estimation }
    \label{fig:c_b_dist}
\end{figure}

Life-saving missions such as Search and Rescue (SAR) operations require professional specialists and state-of-the-art technology to perform life-threatening and physically demanding tasks. The emergence of low-cost, lightweight, and robust UAVs with high payload and long flight time can reduce the danger for SAR specialists by changing the entire operational perspective \cite{vincent_lambert_2023}. State-of-the-art UAVs can be used for tasks such as gathering information for mission planning, real-time mission monitoring, and even execution. Sophisticated UAVs are equipped with GPS and radio communications that can help SAR specialists locate the specific terrain and obtain the information before the operation \cite{polka_2017}. The UAVs equipped with GPS and lightweight vision system can be used for target detection, recognition \cite{illahi_2022}, and robust real-time tracking and following while maintaining a constant safety distance to estimate target's geolocation relative to the UAV \cite{abdelnabi_2024}. Considering the high accuracy and reliability demands of SAR operations, integrating multiple sensors with fusion algorithms, can improve the performance of the UAV tracking and following. The integration of the visual system with depth camera and monocular camera, combined with state-of-the-art deep learning models for detection, recognition, tracking and body keypoint estimation, presents a promising approach to enhance UAV performance under regularly challenging conditions.\\
This paper introduces an innovative framework for camera-to-body (C-B) distance estimation, depicted in Fig. \ref{fig:c_b_dist}. The estimation is based on the depth and monocular camera images enhanced with deep learning models for person detection, recognition, tracking and body keypoint detection. After detecting people and performing face recognition \cite{siktar2025} with You Only Look Once (YOLO) \cite{redmon_2016_CVPR} and Dlib, the system executes tracking using YOLO-based tracking with re-identification \cite{jocher2023yolo}, and a UAV following with C-B distance approximation as the main variable. Deep learning methods robustly achieve detection, recognition, body tracking, and monocular distance approximation \cite{siktar2025}, where tracking involves detecting specific objects and assigning them a specific ID across frames, despite visual challenges \cite{erregue_2025}. The used tracking method that fuses the object detection with re-identification embeddings combines the parallelized YOLOv11 \cite{jocher2024ultralytics} and ByteTrack \cite{zhang2022bytetrack} mechanism to track multiple objects.
The C-B distance approximation utilizes YOLO-pose \cite{jocher2024ultralytics} to extract 2D body keypoints from the monocular camera for distance estimation, as illustrated in Fig. \ref{fig:s_h_distance}. Also, YOLO-pose is employed to acquire depth measurements from the depth camera at the corresponding body keypoint locations. The monocular camera prediction and depth measurement are fused using the EKF to take advantage of both approaches. The fusion of depth and monocular data requires data synchronization, noise management, and real-time deep learning inference to achieve accurate results. The proposed method enables accurate C-B distance estimation under challenging conditions such as reflections and poor visibility by detecting and replacing depth camera outliers  (Algorithm 2)  with monocular camera depth approximation.\\
The main contribution of the proposed work is a EKF-based framework for multi-modal C-B distance estimation, validated against motion capture ground truth data and outdoor tests. The system combines deep learning-based depth approximation from a monocular image with direct depth measurement. The EKF integrates the C-B values derived from the distance measurements of the body keypoints using a monocular image, and the depth measurements for the same body keypoints from the depth camera, as demonstrated in Fig. \ref{fig:s_h_distance}. The body keypoint approach tracks and extracts data across multiple body keypoints invariant to body orientation, surpassing the existing challenges of robustness in body orientation and focusing on primarily face-related keypoints \cite{shen_2018, ollachica_2024}. By fusing the depth camera precision with keypoint-based monocular estimation, robust to reflections and outliers during fast movements, the system extends the depth sensing range while achieveing real-time accuracy, making it ideal for UAV person tracking, following, and broader human-robot interaction tasks.

\section{Related Work}
The vision system for detecting and tracking people, based on an image processing system, offers a new perspective for the planning and execution of SAR operations. Additional visual information from a higher position is useful for the targeted composition and preparation of SAR teams, but also for target localization. The main challenges of using the UAV framework for SAR are the detection, recognition, tracking and following of the target, and the EKF-based sensor fusion algorithm for robust C-B distance estimation.

\subsection{Detection and Recognition}

Some of the most important target detection and recognition techniques for UAVs are deep learning-based approaches. Convolutional neural network (CNN) models such as YOLOv3 \cite{shen_2018}, YOLOv4 \cite{sambolek_2021}, YOLOv9 \cite{hakani2024} for people detection, YOLOv3-tiny and OpenPose for person and body keypoints detection used for gesture recognition \cite{liu2021}, YOLOv8 and YOLOv8-pose for estimating the body position in the given map \cite{henriques2024}. In addition, the Haar cascade classifier \cite{mercado_ravell_2019} for face detection together with the Kalman filter (KF) for tracking the object of interest by estimating the target movements represents an end-to-end approach for face detection and following with the goal of maintaining a constant distance to the face. The distance to the face is measured by the face area, which is compared with the standard face area for male subjects.
Face recognition is performed to match the specific ID of the target person. The most relevant approaches for face recognition are Locality-constrained Linear Coding (LLC) combined with Multi-task Cascaded Convolutional Neural Network for face detection \cite{shen_2018}, OpenFace CNN \cite{balatrusaitis2018}, Dlib face recognition model with Histogram of Oriented Gradients (HoG) as detector \cite{jurevivcius2019method}, FaceNet CNN \cite{schroff2015}, and Siamese recognition model integrated into Simple Matching Real-Time Tracking (SMRT) \cite{ollachica_2024}.

\subsection{Tracking and Following}

The tracking and following procedure pursues the main goal of obtaining the IDs of the detected objects over time. Tracking can be achieved by combining YOLOv3 and Channels and Spatial Reliability Tracking (CSRT), using PD controller for the following procedure \cite{kim_2023}. Computer vision algorithms such as OpenTLD for monocular cameras in combination with the Image Based Visual Servoing (IVBS) controller are often used for tracking \cite{pestana_2014}. The use of the KF is common to estimate the optimal state of the tracked object by fusing the prediction model and the measurements of the real state. The KF can also be used to maintain the tracking of occluded object for a certain period of time \cite{mercado_ravell_2019}. The Tracking Learning Detection (TLD) algorithm uses the Lucas-Kanade algorithm in combination with a detector and a classifier for optimal tracking. The resulting position of the target in the camera image and the distance of the object are then fed into a PID controller to ensure optimal following \cite{bartak_2015}. The state-of-the-art deep learning-based tracker is YOLO11-JDE for joint detection and embedding, which combines YOLOv11 object detection and the Re-ID algorithm for re-identification \cite{erregue_2025}. The approach combining tracking and re-identification mechanisms increases the system robustness to occlusions and image irregularities and is therefore used in our approach.

\subsection{Kalman Filter-based sensor fusion}

UAV frameworks rely on the KF and EKF-based fusion of model predictions and measurements for accurate tracking and following. KF and EKF improve tracking and following tasks by modeling the behavior of the UAV and predicting the next state, which is then corrected by the measurement. It also fuses different sensor data to estimate whether the drone's position \cite{baidya2024enhanced, driessen2018experimentally} or the object position \cite{diaz2018} is correct. Tracking and following can be performed by localizing the face in the image and estimating the depth distance to the face based on the face area. The face area is compared with the reference at known distances. The controller for following the drone keeps the distance constant by combining the model-based distance estimation with real-time detections using KF, meeting the requirements of precision and reliability for human-robot interactions \cite{mercado_ravell_2019}. The SMART-TRACK algorithm combines YOLOv8 object detection and the KF estimator with the depth camera to perform precise tracking. The estimator detects an object and predicts the next object position, which is then fused with the depth camera measurement for correction \cite{gabr2024smart}.

\section{Methodology}

\begin{figure}[h]
    \centering
    \subfloat[\centering]{\includegraphics[width=0.25\textwidth]{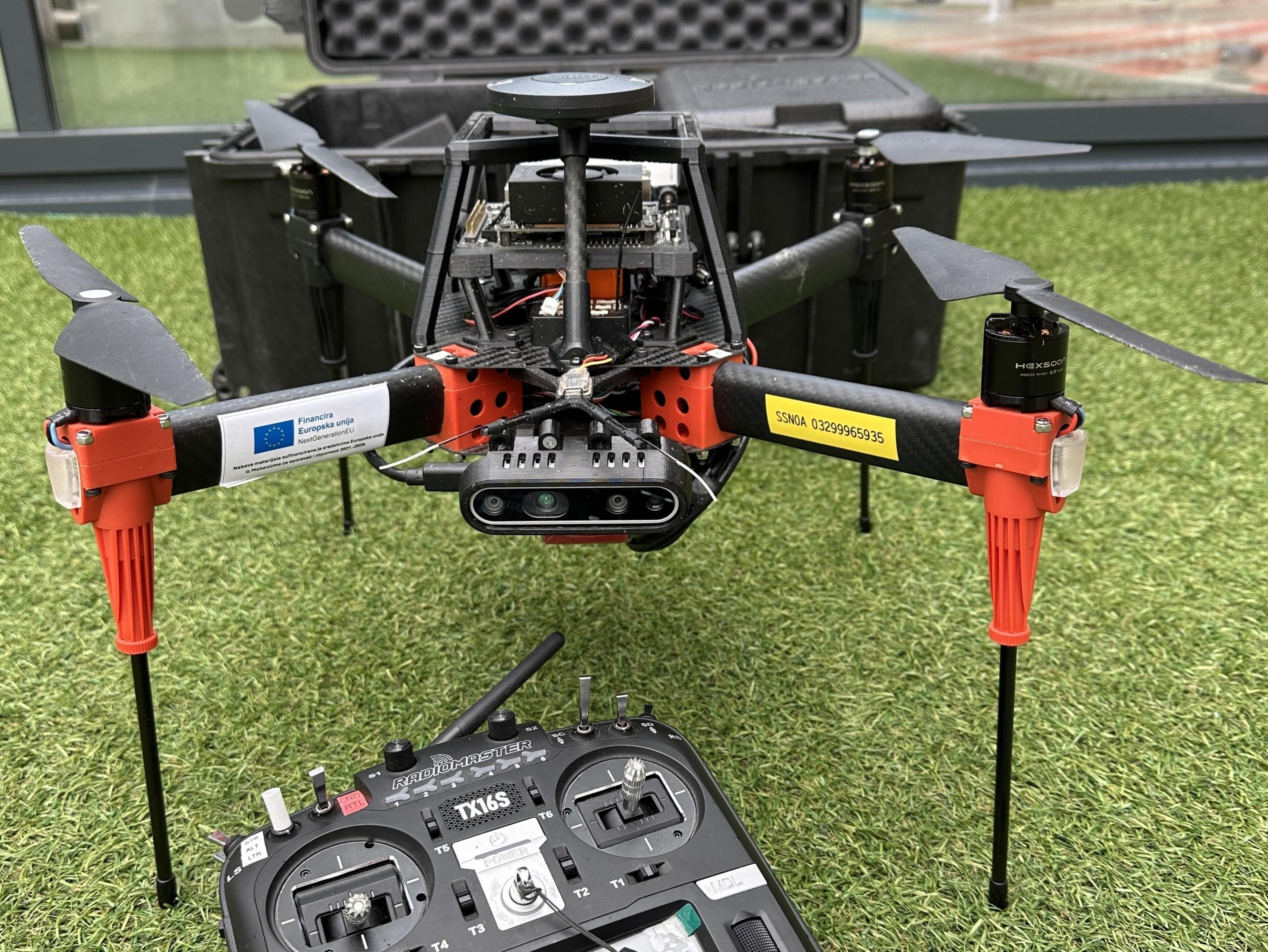}}
    \hfill
    \subfloat[\centering]{\includegraphics[width=0.23\textwidth]{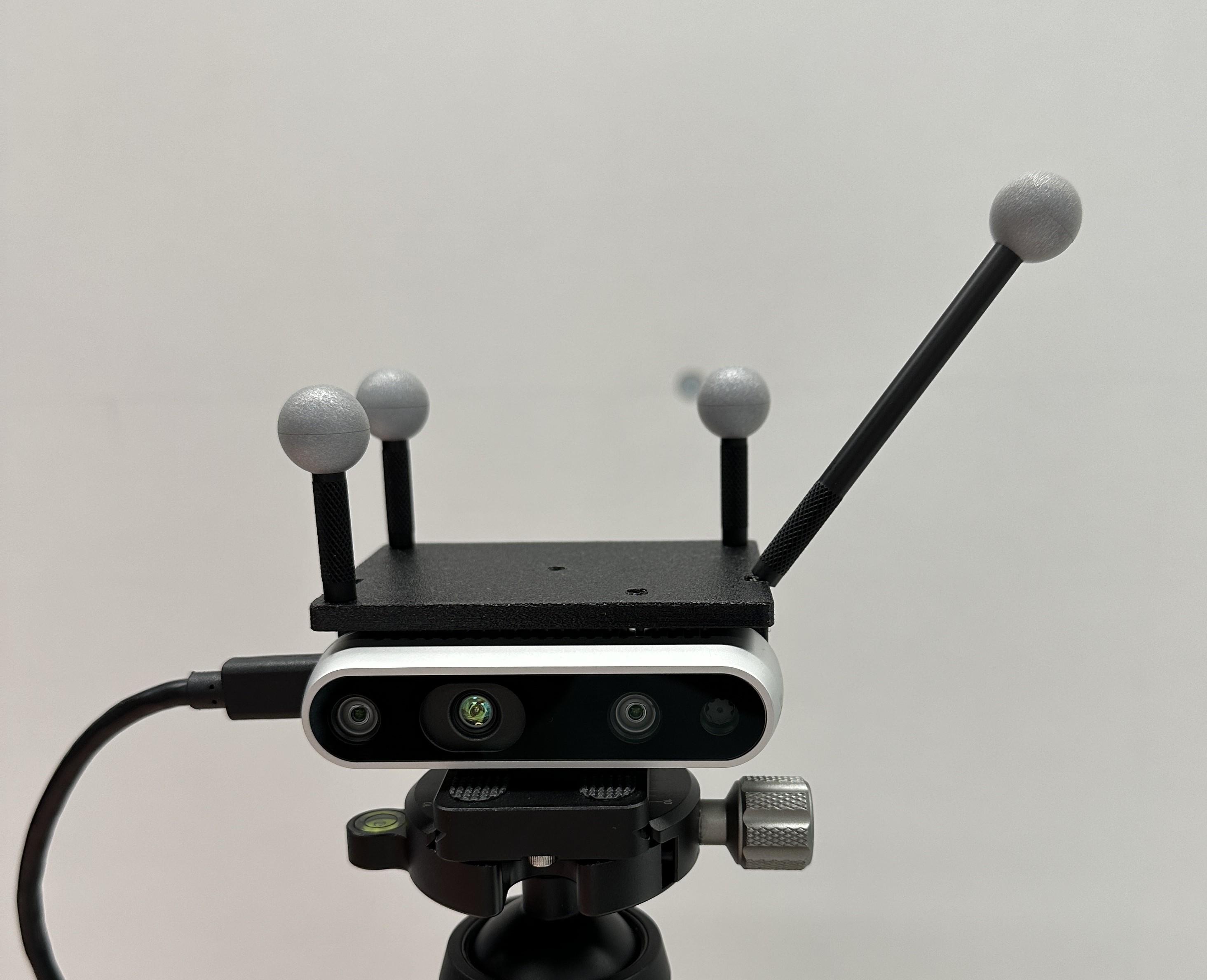}}
    \hfill
    \subfloat[\centering]{\includegraphics[width=0.24\textwidth]{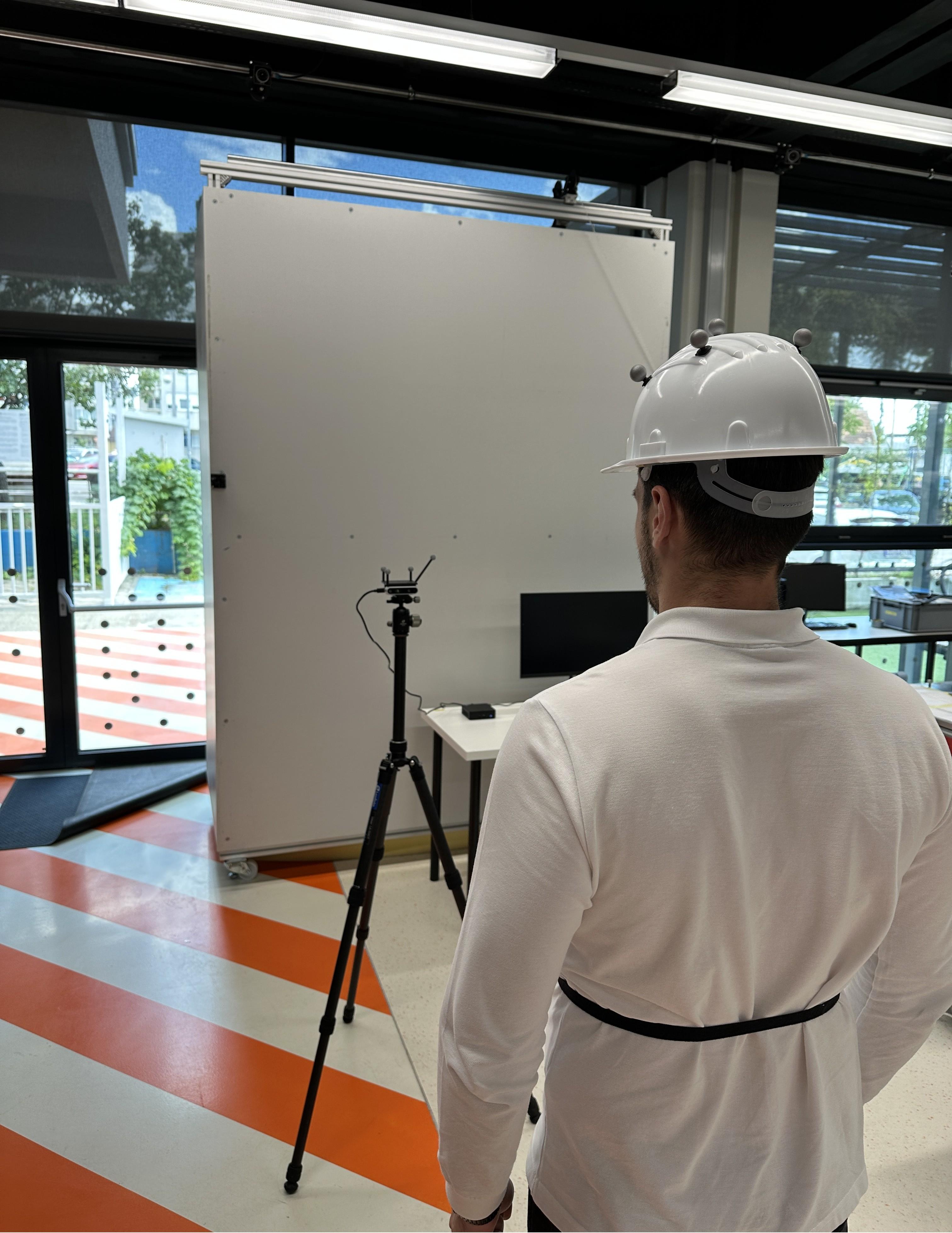}}
    \hfill
    \subfloat[\centering]{\includegraphics[width=0.24\textwidth]{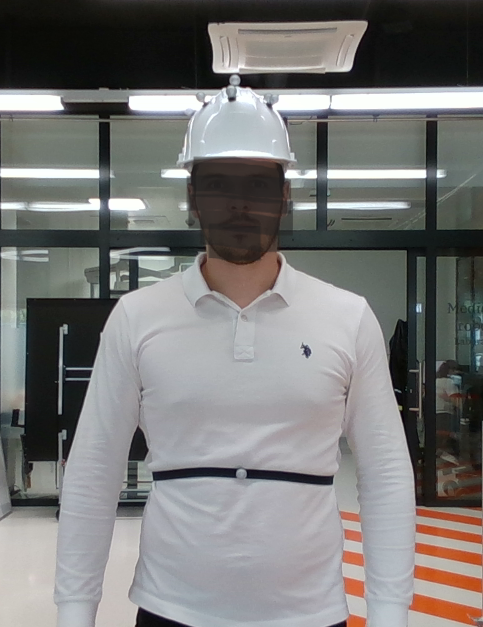}}
    \caption{UAV framework and experimental setup for C-B distance evaluation using OptiTrack. 
    \textbf{(a)} Hexsoon EDU450 with vision system and onboard computer. 
    \textbf{(b)} RealSense and motion tracking markers. 
    \textbf{(c)} and \textbf{(d)}  Person motion tracking markers.}
    \label{fig:drone_markers}
\end{figure}

This paper proposes a EKF-based method for fusing monocular camera depth estimates based on body keypoints and depth camera measurements. The proposed method is part of a visual-based UAV control subsystem that utilizes a PID controller for human-robot interaction when following a person. The visual-based control subsystem utilizes a lightweight onboard computer and mounted vision system, consisting of a depth camera and a monocular camera. The onboard computer enables real-time processing and autonomous mission execution, while the camera enables the fusion of data from both imaging modalities. The target drone for our algorithm is Hexsoon EDU450 with Orange Cube flight controller for on-board communication, shown in Fig. \ref{fig:drone_markers} (a). The on-board computer used is an NVIDIA Jetson Xavier NX equipped with a 6-core NVIDIA Carmel Arm®v8.2 64-bit CPU and a 384-core NVIDIA Volta architecture GPU with 48 tensor cores, suitable for deep learning tasks. The operating system used is Ubuntu 20.04 with the ROS2 package. The deep learning models are provided using Ubuntu 22.04 and PyTorch via Docker. The vision system is an Intel RealSense D435i depth camera with a monocular RGB camera with a resolution of 1920x1080 at 30 frames per second and a depth camera with a resolution of 1280x720 at up to 90 frames per second. The optimum working range for the depth camera extends from 0,4m to 4m. The identical vision system is used for this research and its validation, which focuses on body depth i.e. C-B distance estimation. To evaluate the proposed deep learning based algorithm, we tested the C-B distance estimation process with the OptiTrack motion capture system with 8 cameras in a working area of 40 $m^2$. The UAV system and the experimental setup with markers placed on RealSense camera and human body are shown in Figs. \ref{fig:drone_markers} (b), (c) and (d).

\begin{figure}[h]
    \centering
    \subfloat[\centering]{\includegraphics[width=0.24\textwidth]{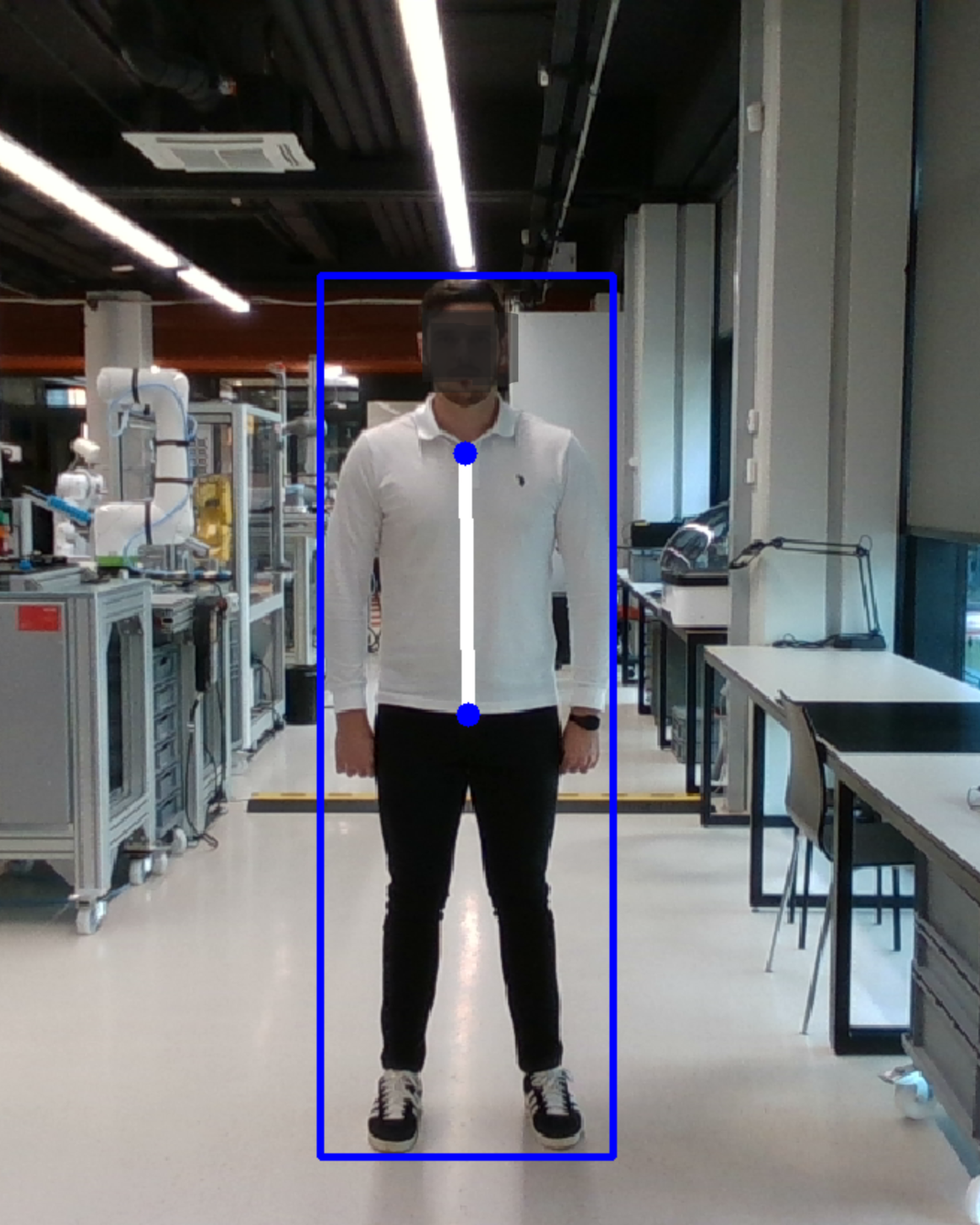}}
    \vspace{0cm}
    \subfloat[\centering]{\includegraphics[width=0.24\textwidth]{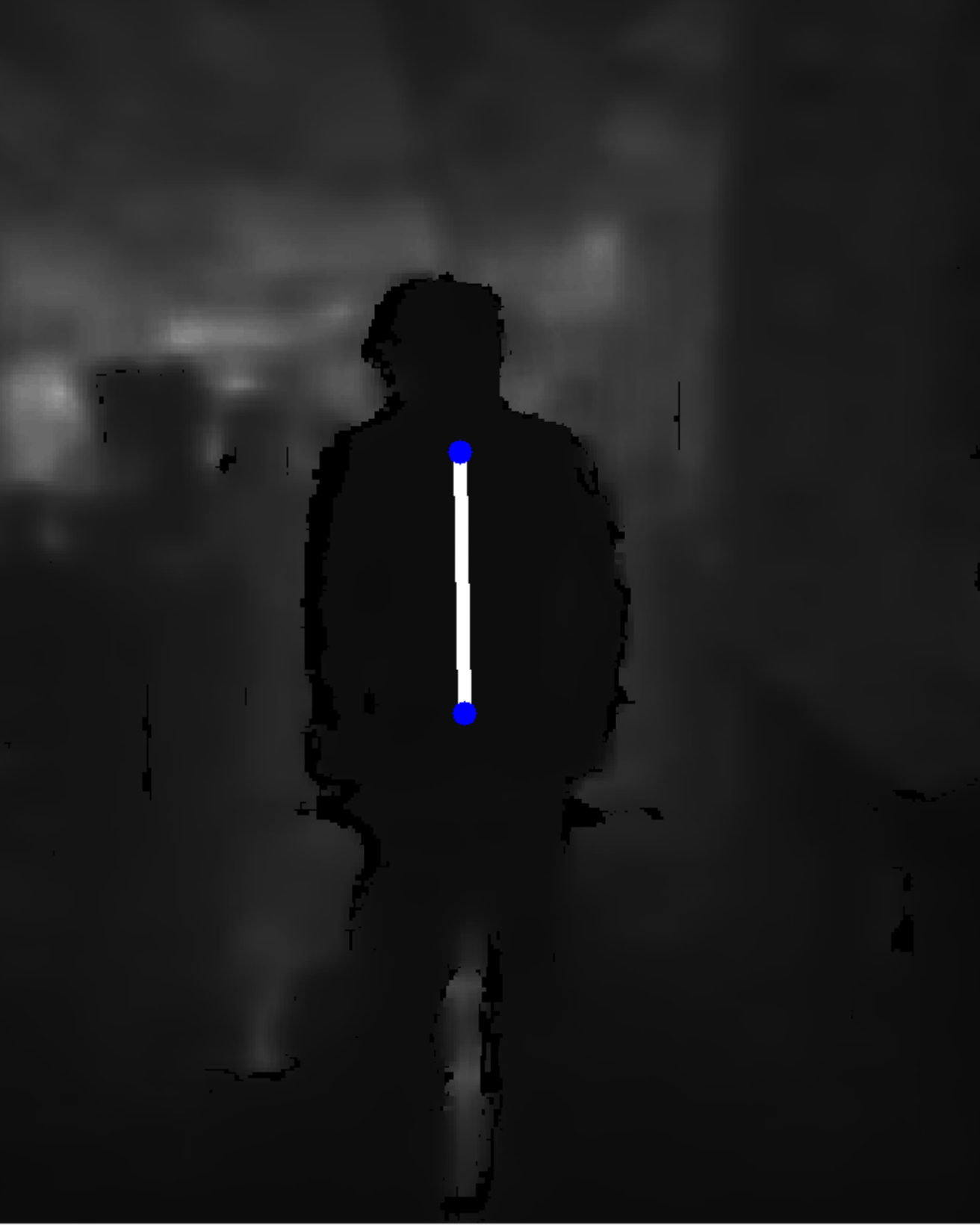}}
    \caption{\textbf{S-H (shoulder-hip) line}. S-H line is used to (a) approximate C-B (camera-to-body) distance from monocular image using eq. \ref{eq:piecewise}, and (b) extract depth information  i.e. C-B distance from the depth camera. 
    \textbf{(a)} S-H line detected on monocular image. 
    \textbf{(b)} S-H line on aligned Depth image.}
    \label{fig:s_h_distance}
\end{figure}

The visual-based control system for UAV target tracking and following integrates two key subsystems. The first subsystem utilizes monocular camera to perform deep learning-based detection, recognition, tracking and following, as described in Algorithm 1 (see Fig. \ref{fig:s_h_distance}) \cite{siktar2025}. The second subsystem proposes using a deep learning-based method for body keypoint detection and their depth data extraction from depth image, suitable for estimating C-B distance, used for following procedure. Person detection is performed with the YOLOv11 CNN model, which enables real-time bounding boxes with high accuracy on edge computer. Based on the detection results, face recognition utilizes Dlib model to identify the targeted person against the provided reference (template) image. If no reference image exists, the system automatically captures the current person detection if it exists, with the identity then confirmed by the UAV operator. Following the recognition step, the system initiates tracking.
\begin{algorithm}
\caption{Visual-Based Control System Workflow}
\begin{algorithmic}[1]
\REQUIRE Monocular camera, YOLOv11 CNN, YOLOv11-pose CNN, Dlib model
\STATE \textbf{Detection}: YOLOv11 
\STATE \textbf{Recognition}:
    \IF{reference image exists}
        \STATE Face recognition using Dlib 
    \ELSE
        \STATE Store current reference image and initiate tracking
    \ENDIF
\STATE \textbf{Tracking}: YOLOv11-pose 
\STATE  Focus on \texttt{shoulder} and \texttt{hip} keypoints to compute S-H distance between their midpoints
\STATE \textbf{Outlier Filtering}: Calculate S-H derivative - distance change
\STATE \textbf{Estimation}: Measure S-H distance and compare to ref. at predefined positions to estimate C-P distance using eq. 1
\end{algorithmic}
\end{algorithm}

Tracking is performed using the YOLOv11 model with re-identification that enables tracked bounding boxes of specific target. The region of interest (inside the tracked bounding box) is extracted from the image and passed to the YOLOv11-pose CNN model for body keypoints detection. The standard model detects 17 body keypoints, with the shoulder and hip keypoints the most important for the proposed algorithm. The distance between the midpoint of the shoulders and the midpoint of the hips (S-H distance) is considered to be the most stable anthropometric dimension for most human body movements and is independent of the orientation of the body to the camera, shown in Fig. \ref{fig:s_h_distance}. The limitation of the method, when the subject leans towards the camera and causes inaccurate approximations, is successfully filtered out as an outlier that is excluded from UAV control to prevent collision. The calculated S-H distance value is filtered by comparing its derivative to the threshold value that represents the derivative of the change when the person moves towards the camera sensor. When using the monocular camera setting in combination with the YOLOv11-pose, the S-H distance can be measured and compared to the reference distance at defined C-B positions. The relationship between the S-H distance measured in pixels and the approximate C-B distance (in centimeters) is non-linear. Considering the average height of a man of 1,80 m, the relationship can be defined as a pair of two non-linear functions for two ranges, eq. 1:

\begin{equation}
f(x) =
\displaystyle
\begin{cases}
-48.03 \ln(x - 179.4) + 401, & \text{if } x < 200 \\[3pt]
-240.2 \ln(x - 47.3) + 1457, & \text{if } x \geq 200
\end{cases}
\label{eq:piecewise}
\end{equation}

Despite the limitation of using the predefined height, which can lead to large approximation errors, this approach successfully preserves the dynamics of a change in approximated distance, which can be used in the EKF as a prediction, to stabilize the output.
As a complement to the deep learning-based monocular approach, we propose a deep learning-based depth camera approach to develop a more robust and accurate distance estimation algorithm. Depth cameras, especially stereo cameras, are widely used for SLAM, obstacle avoidance and accurate 3D tracking of detected objects. Low-cost depth camera systems, such as the Intel RealSense D435i used in this study, provide depth measurements with an accuracy error below 5 mm in the working range (up to 4 m) under optimal illumination. Beyond the optimal range and up to a maximum detection range of 10 m, the error increases exponentially. In addition, sensitivity to reflective object surfaces, insufficient lighting, and the angle between camera and object are the main limitations of depth cameras. These disadvantages lead to measurement outliers, which can be eliminated by fusing depth measurement with depth approximation from first subsystem. As the RealSense D435i depth camera has a horizontal field of view (FOV) of 87° and a vertical FOV of 58°, it is assumed that the depth estimation is noisy and inaccurate at the edges of the FOV. The image overlap at the edges is significantly less accurate than in the center of the FOV, which leads to errors that are registered as outliers.\\
The proposed keypoint-based depth data filtering extracts only distance measurements of a person standing in front of a camera. The system requires the combination of a monocular camera and a depth camera with temporally and spatially aligned images. The subsystem with the monocular camera performs person tracking using YOLOv11 with Re-ID and keypoint detection with YOLOv11-pose as described. From the available body keypoints, the line between the shoulder midpoint and the hip midpoint (S-H line) is constructed, as shown in Fig. \ref{fig:s_h_distance}. The constructed line defines the pixels used by the depth camera to measure the C-B distance. The line is visible in all body orientations relative to the camera sensor, even if the person is leaning towards the camera. The proposed algorithm uses the extracted depth information of the pixels along S-H line and calculates the mean C-B value. The calculated mean value is considered as the true value of the person's distance from the camera sensor. To enable the depth data extraction, both image modalities need to be aligned, monocular image needs to be reshaped to the depth image and have synchronized acquisition to get accurate real-time performance. The proposed depth measurement uses between 50 and 350 depth image pixels along the constructed line for mean value calculation. The camera depth estimation is accurate for the specified optimal working range and overshoots for the tested detection range of up to 7 m, as reported in Fig. \ref{fig:interval1}, which is the main reason behind fusing both modalities.

To improve the quality of depth estimation, we use an EKF, that fuses deep learning-based keypoint detection and C-B distance approximation in the prediction step, with deep learning-based depth camera measurements in the correction step. The C-B distance approximation, modeled as Equation (1), has limited accuracy but remains stable and independent of body orientation, surface reflectance, and the position of the body in the camera’s FOV. On the other hand, depth camera measurements are accurate with occasional oscillations and outliers. With UAVs, especially without a gimbal, sudden camera movements increase the number of measurement outliers. The movements may cause the person to be often at the edge of the depth camera image, resulting in incorrect measurements, so the robust estimation requires reliable outlier suppression. Outlier detection is defined as the derivative of the measured C-B distance. If the deviation between two consecutive samples is greater than 25\% compared to the mean of the deviations in the last 10 samples, then the change is identified as an outlier. If the outliers occur during the depth measurement, the EKF interrupts the correction step and relies only on the model prediction. As soon as no more outliers are detected, the filter continues the correction step. The detailed workflow is shown in Algorithm 2 with EKF algorithm presented in Algorithm 3 \cite{maybeck1990TheKF}.

\begin{algorithm}
\caption{Detailed fusion and outlier rejection workflow}
\begin{algorithmic}
\REQUIRE  Depth camera measurement (C-B distance), Mean C-B distance (last 10 iterations)
\STATE \textbf{Outlier Filtering}: Calculate C-B derivative - \(\frac{d(\text{C-B})}{dt}\)
  \IF{\( |\frac{d(\text{C-B})}{dt}| > 1.25 \times  |\overline{\frac{d(\text{C-B})}{dt}} |\)}
        \STATE \textbf{Outlier detected} - use only keypoint-based S-H to calculate C-B approximation
          \STATE Perform only \textbf{prediction} step (monocular image estimation) - \textbf{equation (\ref{eq:piecewise})}
    \ELSE
        \STATE \textbf{No outlier detected} - use depth camera measurement C-B as correction to keypoint-based  C-B approximation
         \STATE Perform \textbf{full EKF step} - \textbf{equation (\ref{eq:piecewise}) + Algorithm 3}
    \ENDIF
\end{algorithmic}
\end{algorithm}

The EKF operates in two main phases: prediction and correction. 
In our use case, the state vector is defined as
\[
    x_k =
    \begin{bmatrix}
        p_k \\ \dot{p}_k
    \end{bmatrix},
    \]
    where $p_k$ denotes the estimated C-B distance from keypoint-based approximation, defined as Equation (1), and $\dot{p}_k$ denotes C-B derivative.
    We assume a nearly constant velocity model:
    \[
        f(x_{k-1}) =
        \begin{bmatrix}
            p_{k-1} + \Delta t \, \dot{p}_{k-1} \\
            \dot{p}_{k-1}
        \end{bmatrix}.
        \]
        The Jacobian of the process model is:
        \[
            F_k =
            \begin{bmatrix}
                1 & \Delta t \\
                0 & 1
            \end{bmatrix}.
            \]
            The depth camera provides a direct measurement of the C-B distance:
            \[
                h(x_k) = z_k.
                \]
                With the measurement Jacobian:
                \[
H_k =
\begin{bmatrix}
    1 & 0
\end{bmatrix}.
\]
The process noise covariance is chosen as:
\[
    Q_k =
    \begin{bmatrix}
        \sigma_p^2 & 0 \\
        0 & \sigma_{\dot{p}}^2
    \end{bmatrix},
    \]
    where $\sigma_p^2$ = 0.02 controls the uncertainty in the predicted C-B distance and $\sigma_{\dot{p}}^2$ = 0.8 the uncertainty in it's derivative.
    The measurement noise covariance is
    \[
        R_k = \sigma_z^2,
        \]
        where $\sigma_z^2$ = 0.018 denotes the variance of the depth C-B measurement.
        
        \begin{algorithm}
            \caption{Extended Kalman Filter Workflow}
            \begin{algorithmic}[1]
                \STATE \textbf{Initialize:} $\hat{x}_0$, $P_0$
                \FOR{$k = 1,2,\dots$}
                % \Statex \textit{Prediction step}
                \STATE \textit{Prediction step}
                \STATE $\hat{x}_{k|k-1} = f(\hat{x}_{k-1})$
                \STATE $P_{k|k-1} = F_k\,P_{k-1}\,F_k^{T} + Q_k$
                % \Statex \textit{Measurement selection}
                \IF{depth measurement is not an outlier}
                \STATE $z_k = z^{\text{measurement}}_k$
                % \Statex \textit{Correction step}
                \STATE \textit{Correction step}
                \STATE $K_k = P_{k|k-1}\,H_k^{T}\,(H_k\,P_{k|k-1}\,H_k^{T} + R_k)^{-1}$
                \STATE $\hat{x}_k = \hat{x}_{k|k-1} + K_k\bigl(z_k - h(\hat{x}_{k|k-1})\bigr)$
                \STATE $P_k = \bigl(I - K_k H_k\bigr)\,P_{k|k-1}$
                \ENDIF
                \ENDFOR
            \end{algorithmic}
        \end{algorithm}

The Algorithm 3 and specified values suggest the EKF trusts more to the direct measurement from depth camera and uses keypoint depth approximation to stabilize the slight measurement oscillations. With these considerations, the keypoint-based approach can benefit by increasing the accuracy range of the depth camera. The error of the depth camera increases exponentially when the actual distance exceeds 4 m. By using the depth approximation fused with depth measurements, as shown in Figures \ref{fig:interval1}, \ref{fig:interval2}, \ref{fig:interval3}, the depth range can be increased to experimentally tested 7 m. Furthermore, the EKF-based fusion approach successfully filters out the outliers caused by reflections and sudden camera movements, which are common in real UAV applications. The proposed method provides a more robust and accurate C-B distance estimation, which is crucial for maintaining a safe distance during UAV following in SAR operations.

\section{Results}
\begin{figure}[H]
\centering
\begin{minipage}{0.45\textwidth}
    \centering
    \includegraphics[width=\linewidth]{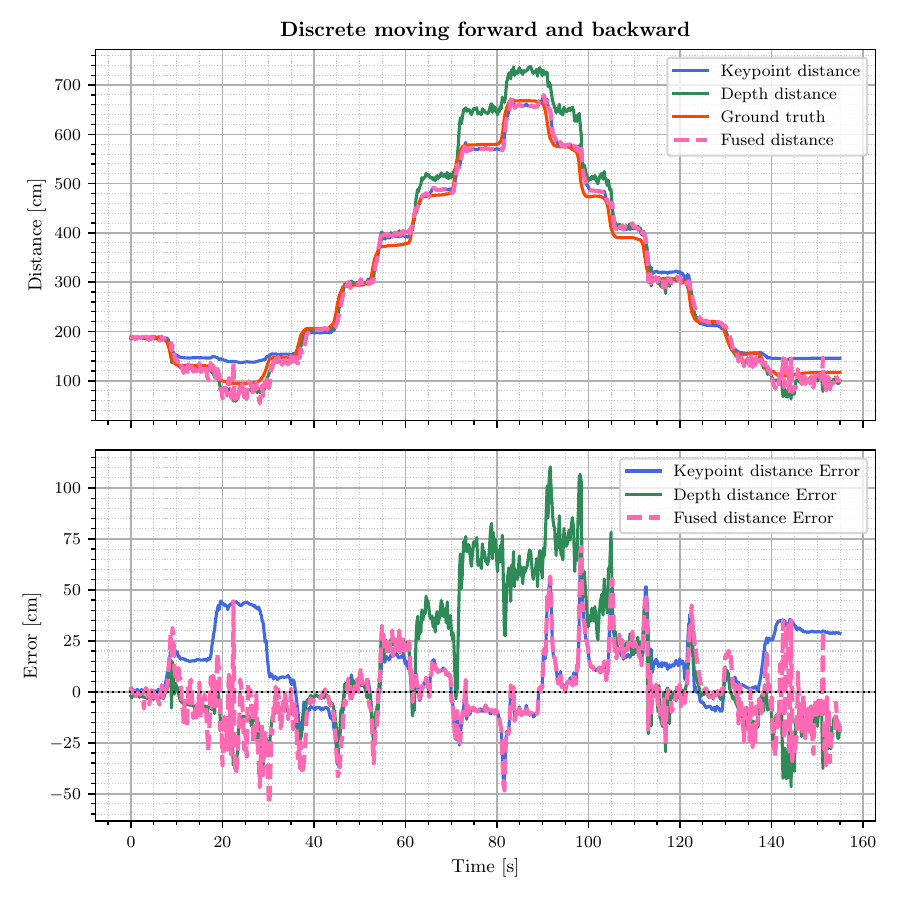}
    \caption{Discrete moving forward and backward: distance and error plots.}
    \label{fig:interval1}
\end{minipage}\hfill
\begin{minipage}{0.48\textwidth}
    \centering
    \captionof{table}{Resulting errors for discrete moving forward and backward}
    \begin{tabular}{lccc}
    \hline
     Method & $|\overline{e}|$ [cm] & RMSE [cm] & $\sigma$ [cm] \\
    \hline
    Keypoint distance & 10.45 & 20.24 & 17.34 \\
    Depth distance    & 13.01 & 34.35 & 31.79 \\
    Fused distance    & \textbf{0.83} & \textbf{17.16} & \textbf{17.14} \\
    \hline
    \end{tabular}
    \label{tab:interval1}
\end{minipage}
\end{figure}

\begin{figure}[!h]
\centering
\begin{minipage}{0.48\textwidth}
    \centering
    \includegraphics[width=\linewidth]{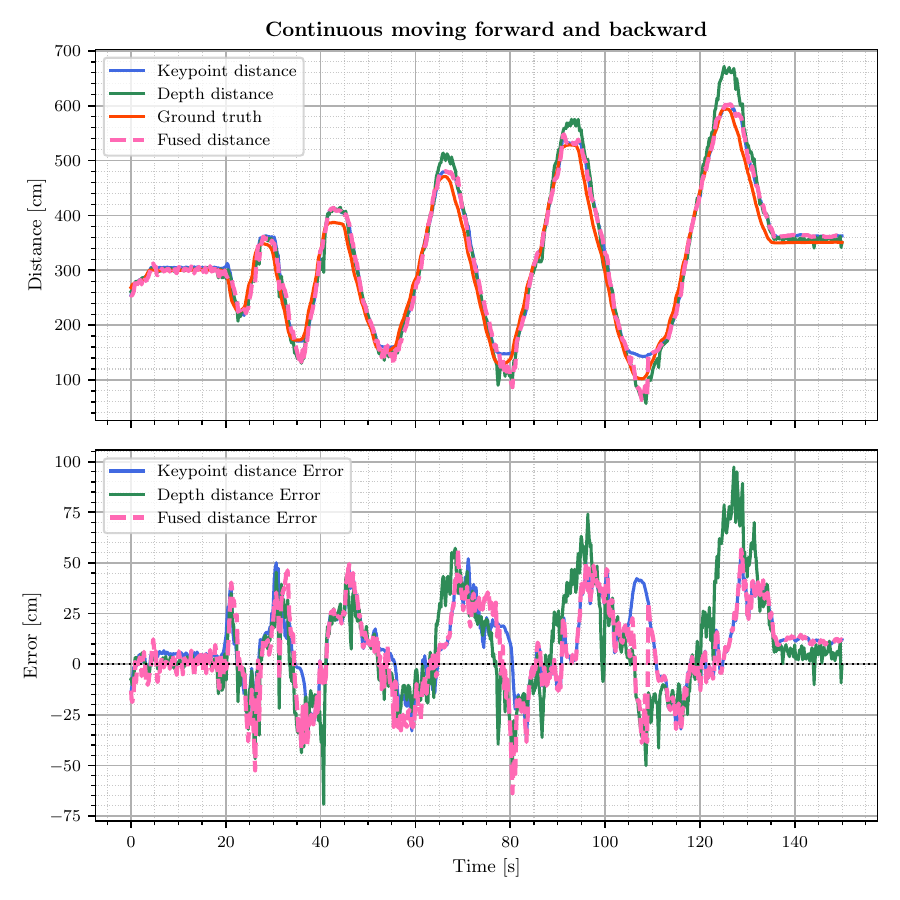}
    \caption{Continuous moving forward and backward: distance and error plots.}
    \label{fig:interval2}
\end{minipage}\hfill
\hspace{0.2\textwidth}%
\begin{minipage}{0.48\textwidth}
    \centering
    \captionof{table}{Resulting errors for continuous moving forward and backward}
    \begin{tabular}{lccc}
    \hline
    Method & $|\overline{e}|$ [cm] & RMSE [cm] & $\sigma$ [cm] \\
    \hline
    Keypoint distance & 9.68 & \textbf{20.41} & \textbf{17.97} \\
    Depth distance    & 8.55  & 26.80 & 25.40 \\
    Fused distance    & \textbf{6.02}  & 21.48 & 20.61 \\
    \hline
    \end{tabular}
    \label{tab:interval2}
\end{minipage}
\end{figure}

\begin{figure}[!h]
\centering
\begin{minipage}{0.48\textwidth}
    \centering
    \includegraphics[width=\linewidth]{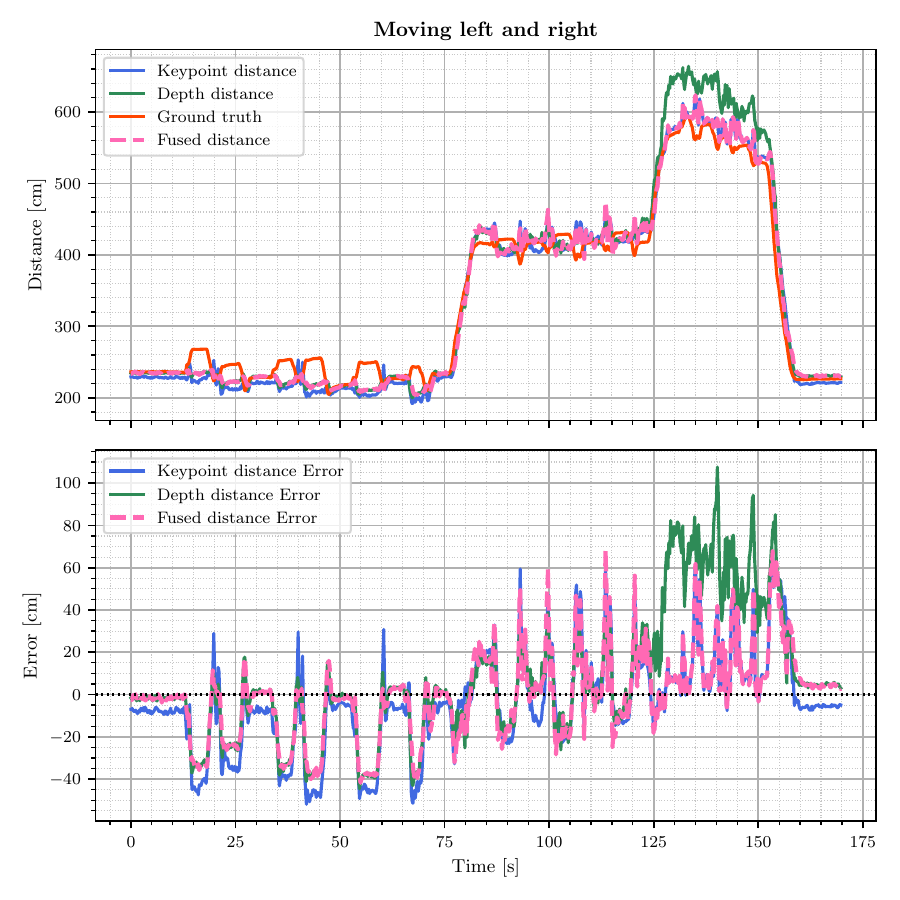}
    \caption{Moving left and right: distance and error plots.}
    \label{fig:interval3}
\end{minipage}\hfill
\begin{minipage}{0.48\textwidth}
    \centering
    \captionof{table}{Resulting errors for moving left and right}
    \begin{tabular}{lccc}
    \hline
     Method & $|\overline{e}|$ [cm] & RMSE [cm] & $\sigma$ [cm] \\
    \hline
    Keypoint distance & 4.09 & 22.90 & 22.53 \\
    Depth distance    & 7.22 & 30.80 & 29.94 \\
    Fused distance    & \textbf{0.61} & \textbf{20.07} & \textbf{20.06} \\
    \hline
    \end{tabular}
    \label{tab:interval3}
\end{minipage}
\end{figure}

The proposed EKF-based fusion method, which combines body keypoint-based depth (C-B distance) approximation with depth camera measurements, is evaluated indoors with static camera, using the OptiTrack motion capture system. Since our system aims to accurately estimate the distance between the camera and the person, we created special marker patterns to define the coordinate systems of the depth camera sensor and the human body. For the RealSense camera, a four-marker pattern was designed, shown in Fig. \ref{fig:drone_markers} b), with the origin at the center of the depth sensor. For the human body, a eight-marker pattern was designed, as shown in Fig. \ref{fig:drone_markers} c). Seven markers were placed on the helmet and one on the abdomen at the midpoint between the shoulders and hips (S-H line), where the origin of the body coordinate frame is defined.
By applying coordinate transformations, the Euclidean distance between the camera coordinate frame and the body coordinate frame is calculated and used as the ground truth for our system.
We evaluated the proposed system in different scenarios: the person walking towards and away from the depth camera step by step with pauses, the person walking continuously in the same way with different amplitudes, and the person walking sideways to the camera and moving left and right to the edges of the depth camera image.
The results of the proposed method are presented in Figs. \ref{fig:interval1}, \ref{fig:interval2} and \ref{fig:interval3}, with the corresponding statistical analysis in Tables I, II and III.

The first scenario, shown in Fig. \ref{fig:interval1} and Table I, involves discrete movements of the person towards and away from the camera. The average distance error for the keypoint-based depth approximation is 10,45 cm, for the depth camera measurements 13.01 cm, while the fused distance error is significantly reduced to 0.83 cm. The RMSE and standard deviation also show a similar trend, with the fused method outperforming both individual methods. The keypoint-based method is not able to estimate the distance when the person is close to the camera due to the limitations of the anthropometric model. On the other hand, the depth camera measurements are more accurate at close range, but they suffer from outliers and noise, especially outside the optimal distance range where the keypoint-based method performs better. The fused method effectively combines the strengths of both methods, resulting in a more robust and accurate distance estimate.

The second scenario, shown in Fig. \ref{fig:interval2} and Table II, involves continuous movements of the person towards and away from the camera. The average distance error for the keypoint-based depth approximation is 9.68 cm, for the depth camera measurements 8.55 cm, while the fused distance error is lower at 6.02 cm. The RMSE and the standard deviation are lower for keypoint-based method, the most stable one for fast body position changes.

The third scenario, shown in Fig. \ref{fig:interval3} and Table III, involves lateral movements, left and right in front of the camera. The average distance error for the keypoint-based depth approximation is 4.09 cm, for the depth camera measurements 7.22 cm, while the fused distance error is 0.61 cm. The RMSE and standard deviation also show a similar trend, with the fused method more accurate than both individual methods. In this scenario, the fused method performs the best, while the depth camera measurements suffer from outliers and noise at the edges of the depth image due to lateral distance movement.

\begin{figure}[!h]
    \centering
    \includegraphics[width=0.48\textwidth]{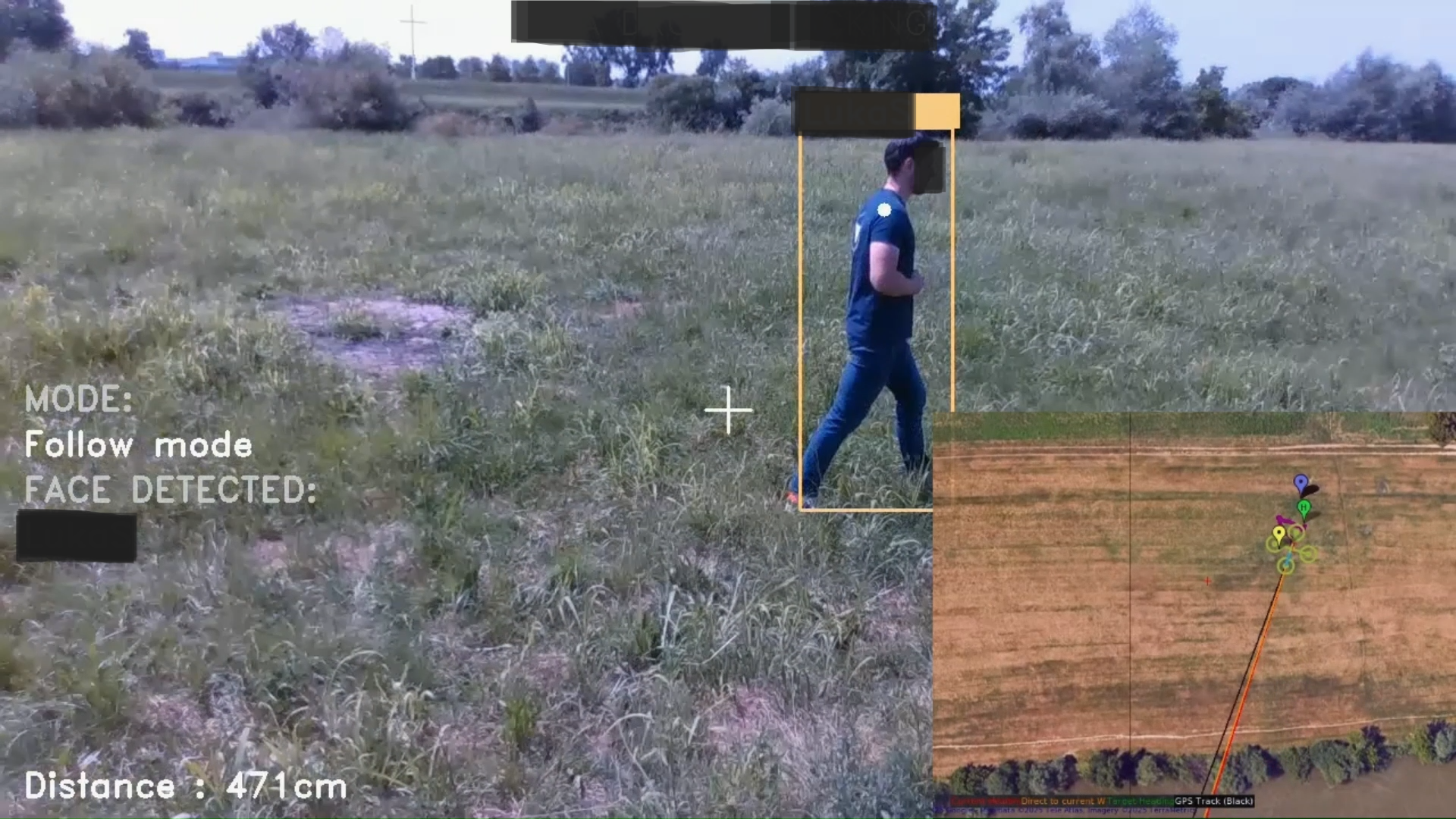}
    \vspace{0cm}
    \includegraphics[width=0.48\textwidth]{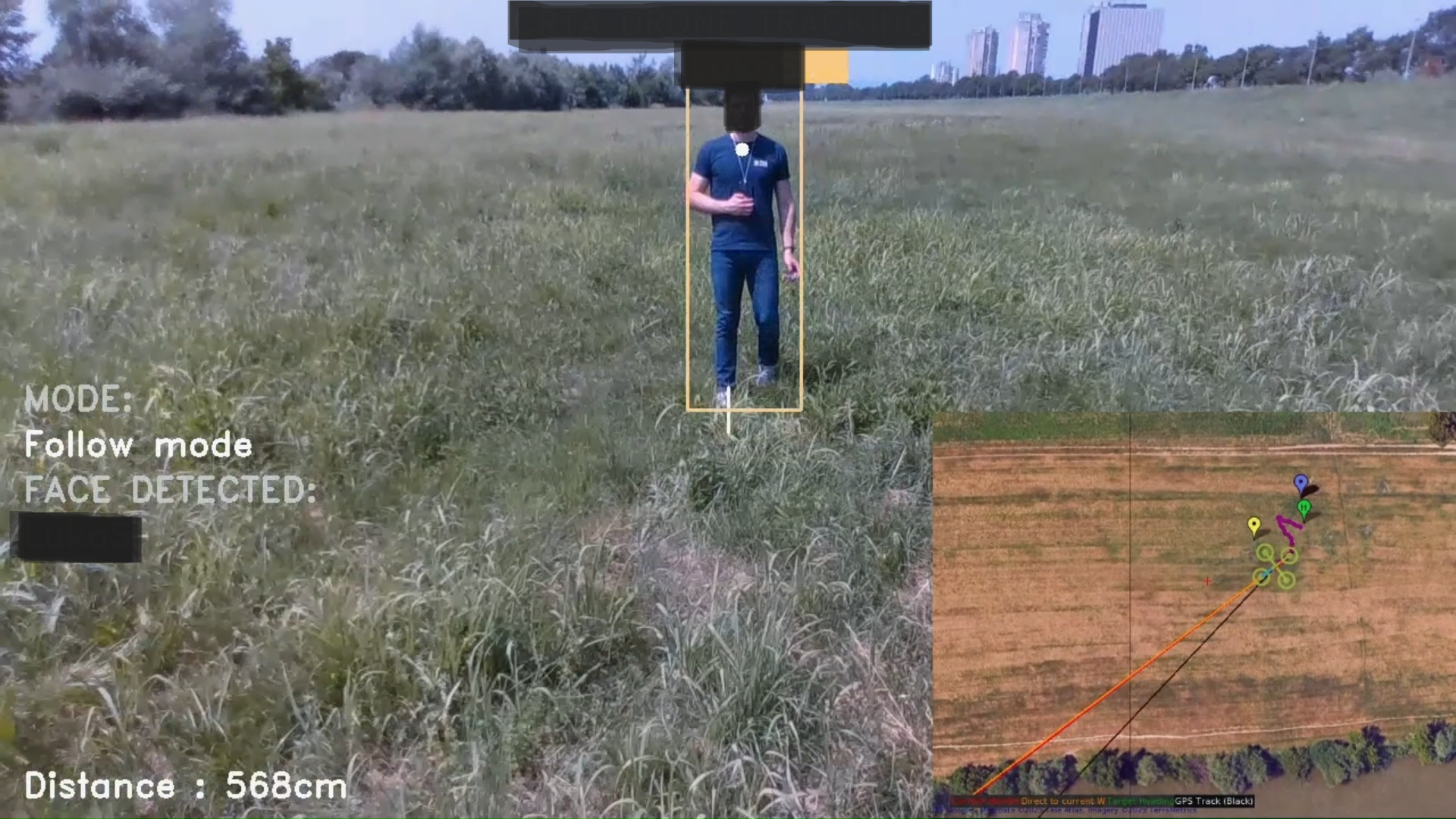}
    \vspace{0cm}
    \includegraphics[width=0.48\textwidth]{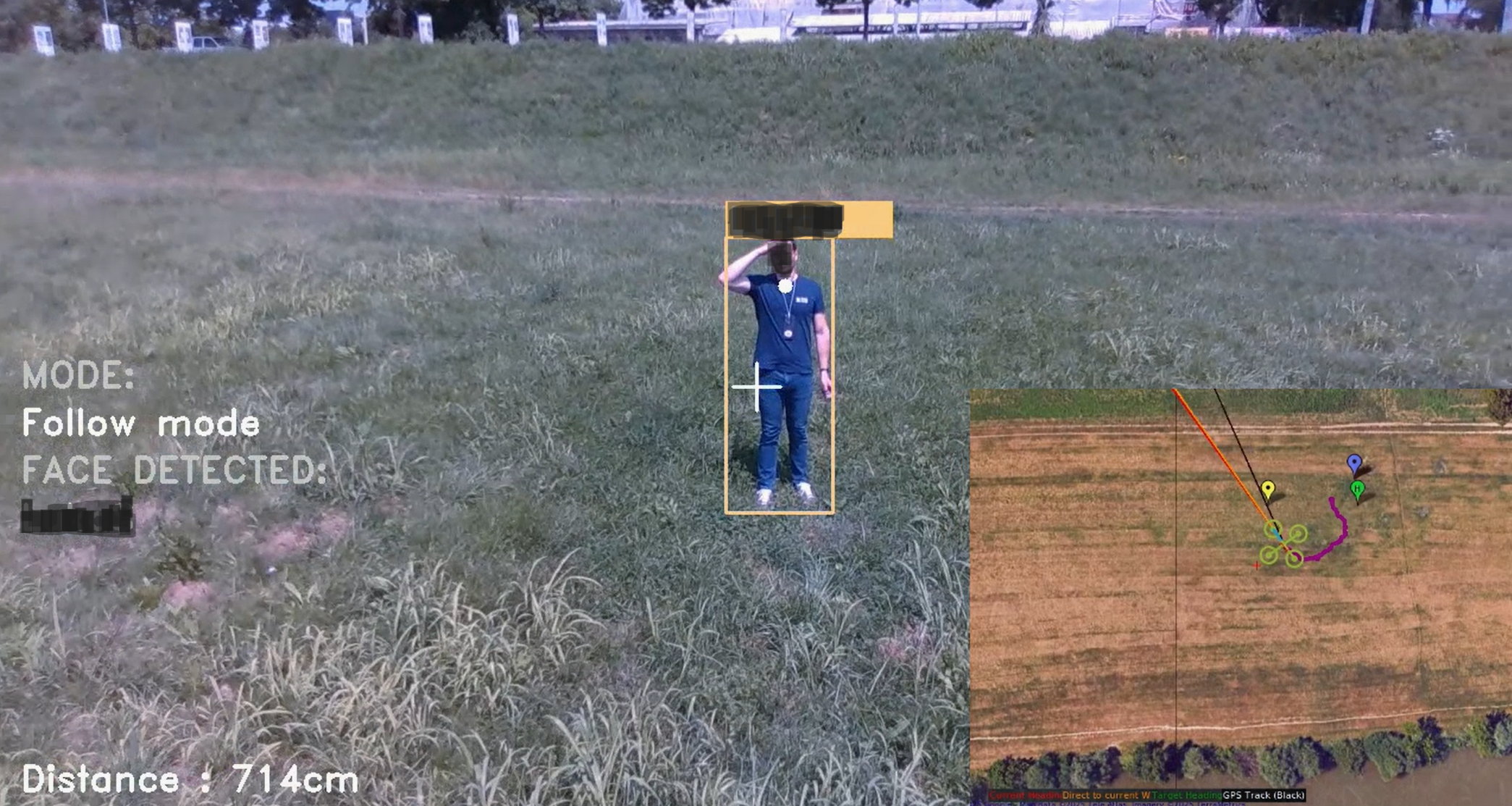}
    \caption{Outdoor evaluation of the proposed fusion algorithm in a restricted environment. \textbf{Bottom Right}: UAV geolocation and person location (yellow pin), with UAV path (purple line). The proposed method successfully tracks the target person in various lighting conditions and eliminates outliers, confirming its suitability for real-time applications in SAR operations.}
    \label{fig:outdoor_results}
\end{figure}

Additionally, the proposed algorithm's performance is evaluated on Hexsoon EDU450 UAV with onboard computer, in outdoor conditions, as shown in Fig.\ref{fig:outdoor_results}. The visual system is evaluated during SAR missions, in which a targeted individual sends an immediate operation request via a GPS-enabled device that communicates with the UAV operating station. The UAV first locates the target using the GPS-provided coordinates. Once the target enters the UAV's visual range, the system performs detection, recognition, and tracking. During the tracking and following procedure the UAV maintains the constant distance to moving target, where GPS map, displayed in Fig. \ref{fig:outdoor_results}, shows UAV location and target location (yellow mark), as well as drone path (purple) during the visual following. The UAV shows robust following by maintaining the safe C-B distance of 5 m, while operating C-B range is set between 3m and 10m, as suggested by indoor results. The outdoor tests in uncontrolled environment, with usual human movements that includes walking and running towards and away from the camera, crouching and laying down, confirm the system's claimed robustness across various lighting conditions, wind disturbances, and GPS inaccuracies. Furthermore, the algorithm operates in real-time at 15 FPS on the Jetson Xavier NX platform, making it suitable as a visual subsystem for SAR applications with autonomous UAVs.

\section{Conclusion}

In this work, we introduce a novel method for robust camera-to-body (C-B) distance estimation. The methodology uses YOLO-pose to detect 2D body keypoints from monocular camera images, computing the shoulder-hip (S-H) midpoint distance to approximate camera-to-body (C-B) distance, while simultaneously extracting corresponding depth measurements from the depth camera at these keypoints. These monocular predictions and depth values are then fused using Extended Kalman Filter (EKF), which manages noise and outliers, and leverages the complementary strengths of both sensors to deliver robust, real-time C-B distance estimation.

Evaluations across indoor and outdoor conditions, including validation against a OptiTrack motion capture system, demonstrate robust performance. Real-time indoor experiments demonstrated up to 15.3\% reductions in mean error, RMSE, and standard deviation across three scenarios. The approach overcomes depth camera limitations such as noise, overshoot, reflections, and poor visibility while extending the optimal sensing range to 7 m. Outdoor tests further confirm its efficacy for real-time UAV applications, including person tracking and following in search-and-rescue (SAR) operations. Depth camera measurements are limited at greater distances and fast movements, whereas keypoint-based distance approximation remains more stable during rapid body position changes, but with a limited accuracy range. To further improve the system's robustness, future work will explore more advanced camera systems and long range depth estimation methods, such as laser-based sensors combined with the proposed keypoint-based approach for automatic target detection. The advanced UAV framework will enable more realistic SAR mission scenarios, that will validate the proposed method in operational conditions.

\section*{ACKNOWLEDGMENT}
%The authors would like to acknowledge the support of the project CRTA Regional Centre of Excellence for Robotic Technology, funded by the ERDF fund and the project MARINERO "Research and development of more innovative products, services and business models in order to strengthen sustainable tourism and green and digital transition of tourism "NPOO.C1.6.R1-I2.01-V3.0011" funded by the NPOO program.
The authors gratefully acknowledge the support of the  VANGUARD (PK.1.1.12.0012) project, funded by the European Union under the Competitiveness and Cohesion Programme 2021–2027.
We would like to acknowledge that Figure \ref{fig:c_b_dist} was created with the help of ChatGPT.

\bibliographystyle{IEEEtran}
\bibliography{references}

\end{document}